\documentclass[runningheads]{llncs}

\usepackage{graphicx}
\usepackage{algorithmic}
\usepackage[ruled,vlined]{algorithm2e} 
\graphicspath{ {./images/} } 

\begin{document}
\title{Human-in-the-loop model explanation via verbatim boundary identification in generated neighborhoods}
%
%
\author{Xianlong Zeng \and
Fanghao Song \and
Zhongen Li \and 
Krerkkiat Chusap \and
Chang Liu}

\authorrunning{Zeng et al.}
%
\institute{School of Electrical Engineering and Computer Engineering, Ohio University, Athens, OH 45701 USA \\
\email{\{xz926813, liuc\}@ohio.edu}}
\maketitle              
\begin{abstract}
The black-box nature of machine learning models limits their use in case-critical applications, raising faithful and ethical concerns that lead to trust crises. One possible way to mitigate this issue is to understand how a (mispredicted) decision is carved out from the decision boundary. This paper presents a human-in-the-loop approach to explain machine learning models using verbatim neighborhood manifestation. Contrary to most of the current eXplainable Artificial Intelligence (XAI) systems, which provide hit-or-miss approximate explanations, our approach generates the local decision boundary of the given instance and enables human intelligence to conclude the model behavior.  Our method can be divided into three stages: 1) a neighborhood generation stage, which generates instances based on the given sample; 2) a classification stage, which yields classifications on the generated instances to carve out the local decision boundary and delineate the model behavior; and 3) a human-in-the-loop stage, which involves human to refine and explore the neighborhood of interest. In the generation stage, a generative model is used to generate the plausible synthetic neighbors around the given instance. After the classification stage, the classified neighbor instances provide a multifaceted understanding of the model behavior. Three intervention points are provided in the human-in-the-loop stage, enabling humans to leverage their own intelligence to interpret the model behavior. Several experiments on two datasets are conducted, and the experimental results demonstrate the potential of our proposed approach for boosting human understanding of the complex machine learning model.

\end{abstract}
\section{Introduction}
Machine learning models are typically designed and fine-tuned for optimal accuracy, which often results in layers of weights that are difficult to explain or understand. In the meantime, recent successes of machine learning systems have attracted adoption from more end-users, who need to better understand the model in order to trust or properly use such machine learning systems. To make these two ends meet, researchers and practitioners alike have adopted several approaches, including 1) using approximate models just for explanation\cite{ancona2019explaining}; 2) linear local explanation for complex global models (e.g. LIME\cite{lime}); 3) example-based explanation by finding and showing most influential training data points\cite{kabra2015understanding}. These approaches all have their own merits, but none of them deliver everything needed by end-users\cite{rudin2019stop}. 

The fundamental limitation of these approaches is that they assume that 1) certain aspects of machine learning systems, especially complex deep neural networks, cannot be understood by human beings, and 2) typical human users can only understand simple concepts such as linear systems.

We have an opportunity to improve on previous attempts with two assumptions. First, human users are intelligent, just not in the same way as machines. Humans can identify patterns intelligently but may not be able to scale up to thousands of data points easily. Second, machine learning systems are built to reflect actual physical systems that follow logical and physical rules. What worked well most likely can be explained, even though the explanation could be complex. What cannot be explained most likely is not a good reflection of the underlying physical properties.

We intend to make improvements in this area by 1) presenting various aspects of the actual model through verbatim model manifestation (instead of trying to approximate the models), and 2) identifying and generating a manageable number of data points to present to users in the local context of the point-of-interest, so that human users can use their own intelligence to understand what the actual model is trying to do within a limited scope that is manageable by a human being.

With this intuition, we aim to design an approach to facilitate human users' understanding of machine learning models through 1) verbatim manifestation of certain aspects of the underlying machine learning systems and 2) contextualized visualization of carefully curated or generated data points that facilitates human understanding. In other words, we try to build a bridge between machine and human intelligence to address machine learning models' explainability problems. Furthermore, we observe that a typical human user does not need to understand the complete machine learning model to gain confidence in the results from the model. The user only needs to understand the rationale behind the decision related to the current task.

\begin{figure}[h]
\centering
\includegraphics[width=0.95\textwidth]{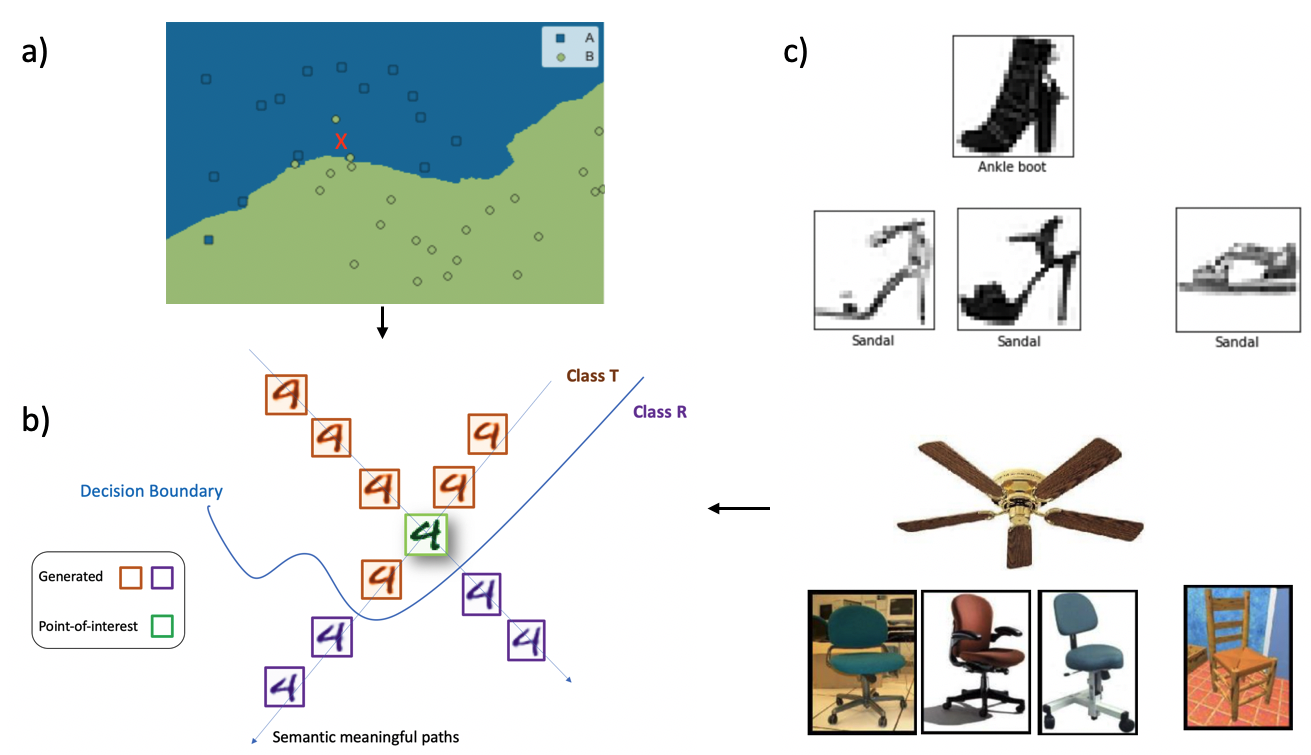}
\caption{A high-level illustration of our proposed framework. a) For a (mispredicted) point-of-interest (red x) and a trained machine learning model, b) our framework tries to carve out the local decision boundary and delineate the model behavior through a manageable neighborhood manifestation. c) Images of \textit{sandals} and \textit{ankle boot} from the fashionMNIST dataset that cause confusion to a classifier. Human users can understand the classification errors by seeing the context that some sandals have boot-shape heels. Another classification error is from the Caltech 101 dataset. Trust crisis can be mitigated given the context that some chairs have fan-shaped bases.}
\label{idea}
\end{figure}

In this paper, we present a three-stage human-in-the-loop XAI system, a high-level illustration of which is depicted in Figure~\ref{idea}. For a given (mispredicted) point-of-interest, our framework tries to carve out its local decision boundary and delineate the model behavior through a neighborhood manifestation. Our framework leverages variational autoencoders (VAE) to generate neighborhood examples that cross the decision boundary. Human users are involved in exploring the neighborhood through three carefully designed intervention points. These intervention points help human users limit the neighborhood's scope and enable them to gain insights from the model behavior. The source code of our work is public available on GitHub: \url{https://github.com/drchangliu/xai}.  

The main contributions of our work are:
\begin{itemize}
  \item We proposed a novel human-in-the-loop framework that could mitigate the trust crisis between human users and machine learning models. 
  \item Several case studies are presented to illustrate the potential of our approach to facilitating human understanding of complex machine learning models.
  \item A general framework to depict the local decision boundary around the (mispredicted) instance-of-interest.
\end{itemize}

\section{Related Work}
Machine learning researchers and practitioners have always used techniques and tools to better understand machine learning models. In this section, we examine a few state-of-the-art tools that are publicly accessible in an attempt to shed some light on how they can help software engineers adopt machine learning components.

To understand the information flow of a deep network, Ancona et al.\cite{ancona2017towards} has studied the problem of assigning contributions to each input feature of a network. Such methods are known as \textit{attribution methods}, which can be divided into two categories: perturbation-based and backpropagation-based. The perturbation-based methods, such as Occlusion \cite{zeiler2014visualizing}, LIME \cite{lime} and Shapely value \cite{ancona2019explaining}, change the input features and measure the difference between the new output and the original output, while backpropagation-based methods compute the attributions for all input features through the network. Backpropagation-based methods include the feature-wise manner and the layer-wise manner. Feature-wise approaches includes Gradient*Input \cite{shrikumar2016not} and Integrated Gradients  \cite{sundararajan2017axiomatic}). Layer-wise approaches includes Layer-wise Relevance Propagation \cite{bach2015pixel}, Class activation maps \cite{Simonyan2014DeepIC}\cite{selvaraju2017grad}\cite{chattopadhay2018grad}\cite{wang2020score} and DeepLIFT \cite{shrikumar2017learning}.

Among these related research efforts, LIME \cite{lime} and DEEPVID \cite{wang2019deepvid} are the two most relevant methods as compared to our framework. LIME, proposed by Ribeiro et al., was an approach that was able to explain the predictions of any model\cite{lime}. LIME utilized a locally interpretable model to interpret the black-box model's prediction results and constructed the relationship between the local sample features and the prediction results. Explanations from LIME do not exactly reflect the underlying model. LIME describes the prediction outcomes obtained even with different complex models, such as Random Forest, Support Vector Machine, Bagged Trees, or Naive Bayes. LIME can handle different input data types, including tabular data, image data, or text data.

DEEPVID, proposed by Wang et al., was a visual analytics system that leverages knowledge distillation and generative modeling to generate a visual interpretation for image classifiers \cite{wang2019deepvid}. Given an image of interest, DEEPVID applied a generative model to generate samples near it. These generated samples were used to train a local interpretable model to explain how the original model makes the decision. The difference between our approach and DEEPVID is that, instead of utilizing interpretable models such as linear regression to provide interpretation, our approach visualizes boundary examples directly. End-users can then leverage their human intelligence to interpret the model decision.

DeepDIG \cite{karimi2019characterizing}\cite{karimi2020decision}, developed by Karmi et al, was a framework that used to characterize the decision boundary for deep neural networks. The main contribution can be divided into two parts. The first part is to generating borderline instances that are near the decision boundary. This part is completed in three steps, the first and second steps are used to generate adversarial instances by Autoencoder. The third step is used to generate the borderline instances based on the binary search and adversarial instances produced after step one and step two. The second contribution is related to the characterization that is used to measure the decision boundary complexity in the input space and embedding space. The input space complexity is calculated by the generated borderline instances from the first contribution. The embedding space complexity is measured by developing a linear Support Vector Machine (SVM) model.

\section{The proposed human-in-the-loop framework}
Given a trained machine learning model and a (mispredicted) point-of-interest, we intend to generate a neighborhood that can enable a better human understanding of the model. The generated neighborhood needs to satisfy three critical criteria:

\begin{itemize}
  \item The instances in the neighborhood need to be semantically close to the point-of-interest.
  \item The decision boundary is at least partially visible within the neighborhood.
  \item The neighborhood needs to maintain the number of instances in a manageable size so that human users can gain insight from it.
\end{itemize}

\begin{figure}[h]
\centering
\includegraphics[width=0.95\textwidth]{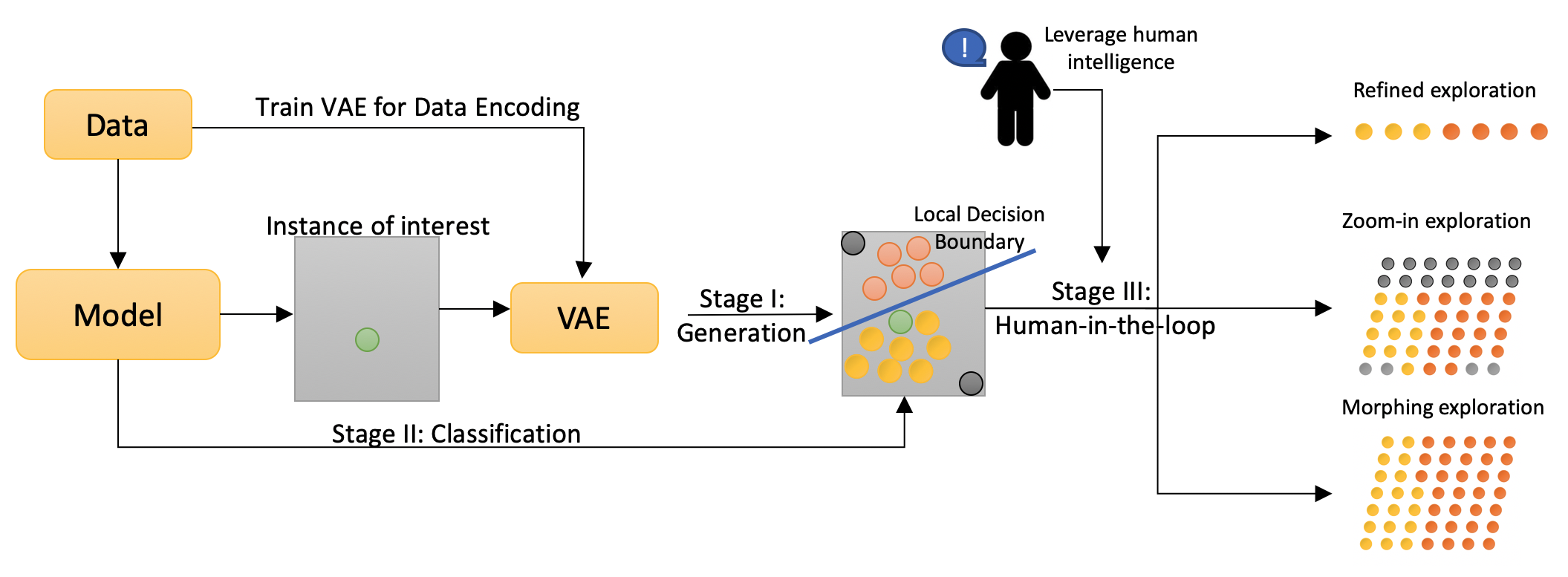}
\caption{The proposed human-in-the-loop framework. It contains three stages. In stage (I), a neighborhood is generated based on the given sample through a trained variational autoencoder. In stage (II), the pre-trained machine learning model is used to yield classification on the generated instances to carve out the local decision boundary and delineate the model behavior. In stage (III), human users are enabled with three intervention points to explore the neighborhood: a) refined multifacet path exploration, b) ``zoom-in''\& ``zoom-out'' area exploration, and c) boundary-crossing morphing exploration.}
\label{pipeline}
\end{figure}

To generate a neighborhood that can satisfy the above three criteria, we propose the human-in-the-loop framework that contains three stages, as shown in Figure~\ref{pipeline}. In the first stage, a neighborhood is generated based on the given sample through a trained generative model. In the second stage, the pre-trained machine learning model is used to yield classification on the generated instances to carve out the local decision boundary and delineate the model behavior. Next, three intervention points are provided to enable human users for a throughout exploration for gaining insights. In the following section, we explain each stage in detail.

\subsection{Stage (I): Neighborhood Generation}
Stage one can be described as a stochastic process that generates neighbors from the given point-of-interest. There are two approaches to accomplish such a procedure: Variational Auto-Encoders (VAEs) and Generative Adversarial Networks (GANs). Both of these two generative methods assume an underlying latent space that is mapped to the original data space through a deterministic parameterized function. The generative model often consists of an encoder that can map the given data into the latent space, and a decoder that can decode the latent space vector back to the original space. In this work, we adopt VAE as the generative model because of its more straightforward model structure. 

\begin{figure}[h]
\centering
\includegraphics[width=0.95\textwidth]{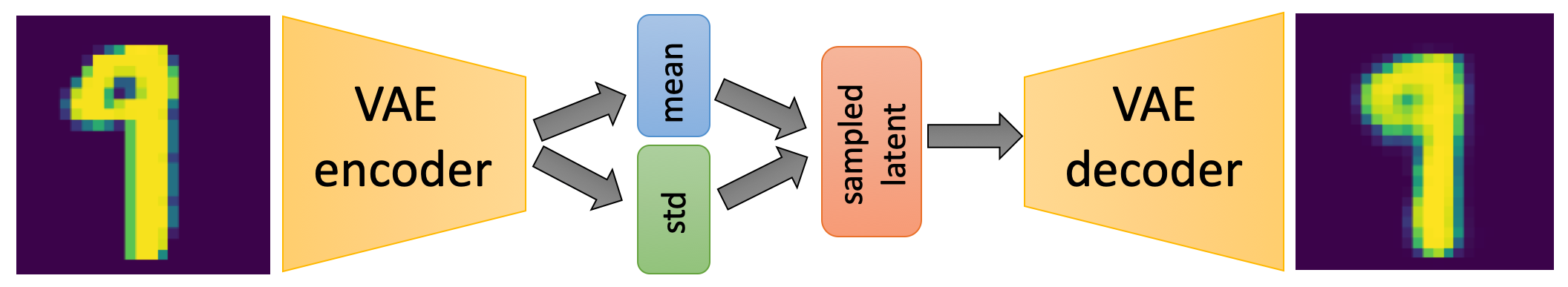}
\caption{The architecture of our selected generative model, i.e., a Variational AutoEncoder (VAE)}
\label{VAE}
\end{figure}

As shown in Figure~\ref{VAE}, we train an encoder-decoder CNN-VAE with ten latent dimensions on the MNIST dataset to learn the underlying latent distribution. A hyper-parameter \textit{step-length} is applied to each latent space via linear interpolation to generate the perturbed latent vectors. The perturbed latent vectors are then fed through the decoder to generate neighbors around the point-of-interest. 

More formally, a VAE model that consists of encoder $q_\theta(z|x)$ and decoder $q_\phi(x|z)$ are trained on the dataset X, where $X = \{(x_1,y_1), (x_2,y_2), ..., (x_n,y_n)\}$, $x_i \in R^D$ and $y_i\in[1,c]$. The VAE is trained with the negative log-likelihood with regularizer. The loss function $l_i$ for data instance $x_i$ is:

\begin{equation}
-E_{z-q_\theta{(z|x_i)}}[log_{p_\theta}(x_i|z)] +KL(q_\theta(z|x_i)||p(z)),
\end{equation}

where $z\in R^d$ denotes the \textit{d-dimension} embedding space learned by the VAE encoder.

Utilized by the trained VAE, examples near the point-of-interest can be generated and form the neighborhood. A hyper-parameter \textit{step-length} needs to be chosen to determine the border of the neighborhood. In practice, we set \textit{step-length} equal to one as the default value.

\subsection{Stage (II): Neighborhood Classification}
To identify and visualize the local decision boundary, the given trained machine learning model is applied to the generated instances. The classification results are highlighted with different colors so that the model behavior can be delineated. We call this classification results as classified neighborhood. A classified neighborhood is one where every data point within the neighborhood has been classified by the model-under-investigation so that the decision boundary is identified and visualized verbatim. Because the actual model is used, this is a verbatim manifestation of the model decision boundary within the neighbourhood. In practice, a larger value of \textit{step-length} is recommended to ensure a decision boundary with clear difference between the opposite sides. In our experiments, we set the \textit{step-length} to 1. 

\subsection{Stage (III): Human-in-the-loop Exploration}
Three intervention points are provided in our human-in-the-loop stage. Specifically,
\begin{itemize}
  \item a refinement intervention point that provides a multifacet refined neighborhood exploration. 
  \item a ``zoom-in" \& ``zoom-out" intervention point that enables human users to take a closer look at the certain region of interest.
  \item a morphing intervention point that selects two examples from each side of the decision boundary and creates a visualization path.
\end{itemize}

For the first intervention point, human users are enabled to identify the dimensions of interest, i.e., specific dimensions from the d-dimensional latent space. Next, we allow the human to adjust the hyper-parameter \textit{step-length} along the selected latent dimension for exploration. A larger value of the \textit{step-length} will enrich the semantic variation, while a smaller value can provide a more concentrated result. The \textit{step-length} serves as a "tuning knob" to adjust traversal speed in the latent space, which helps human users to understand how a prediction is carved out from specific changes.

Human users are allowed to identify two hidden dimensions of interest for the second intervention point and construct a morphing matrix based on these two-dimension spaces. Allowing the morphing of two dimensions simultaneously can provide a richer context around the point-of-interest. The second intervention point acts as a ``zoom-in''\& ``zoom-out'' effect to assist human users in gathering insights from the generated examples.

For the third intervention point, a few instances that are semantically close to the given point-of-interest at two sides of the decision boundary are provided. Next, a morphing path between the two instances are created and the path passes through the point-of-interest. The algorithm for identifying the nearest neighbor and creating the morphing path is shown in Algorithm 1. Such morphing traverses data manifold while crossing the decision boundary, which can delineate the model behavior and explain how and why a particular image is relevant to the prediction.

\begin{algorithm}[h]
\SetAlgoLined
\textbf{Given:} Dataset $(X,Y)$\\ 
\textbf{Given:} Classifier $F()$ to be interpreted\\
\textbf{Given:} Pretrained \emph{VAE: (VAE-enc, VAE-dec)}\\
\textbf{Given:} Data instance of interest $(x_i, y_i)$, where $y_i=c_1$, but mispredicted $F(x_i)=c_2$\\
\begin{algorithmic}[1]
    \STATE $\emph{enc-}x_i = \emph{VAE-enc}(x_i)$
    \FOR{{$(x_j,y_j) \in (X,Y), y_j=c_1$}}
        \STATE $\emph{enc-}x_j = \emph{VAE-enc}(x_j)$
        \STATE update $x_j$ s.t. $\|\emph{enc-}x_j - \emph{enc-}x_i\|_{L1}$ is smallest
    \ENDFOR
    \FOR{{$(x_k,y_k) \in (X,Y), y_k=c_2$}}
        \STATE \emph{enc-}$x_k =$ \emph{VAE-enc}$(x_k)$
        \STATE update $x_k$ s.t. $\|\emph{enc-}x_k - \emph{enc-}x_i\|_{L1}$ is smallest
    \ENDFOR
    \STATE interval=($\emph{enc-}x_k - \emph{enc-}x_i)$/num-neighbors 
    \STATE neighbors=[]
    \STATE labels=[]
    \FOR{{i=0, i$\leq$num-neighbors; i++}}
        \STATE neigh = $\emph{enc-}x_i \pm $interval 
        \STATE neighbors.append(neigh)
        \STATE labels.append($F$(neigh))
    \ENDFOR
    \STATE Visualize(neighbors, labels)
\end{algorithmic}
 \caption{Pseudocode for the proposed method}
\end{algorithm}

\section{Experiment Setup}
To verify the effectiveness of our proposed framework, we conduct several experiments on two datasets. Section 4.1 describes the datasets and the trained machine learning model architectures. Section 4.2 presents the detailed experimental settings for our framework.

\subsection{Dataset and Trained machine learning Architecture}
We investigate the proposed framework against two datasets, MNIST and FashionMNIST. The MNIST dataset is a large database of handwritten digits, while FashionMNIST is a dataset of Zalando's article images. The images in these datasets are 28x28 grayscale images associated with a label of 10 classes. Both MNIST and FasionMNIST are commonly used for training various image processing machine learning models. The details of the datasets and the chosen model performance are shown in Table~\ref{tab1}.

\begin{table}
\caption{Description of the investigated datasets.}\label{tab1}
\begin{tabular}{|l|l|l|}
\hline
 &  MNIST & FashionMNIST\\
\hline
\# of training examples &  60,000 & 60,000\\
\hline
\# of testing examples &  10,000 & 10,000\\
\hline
\# of output classes & 10 & 10\\
\hline
Original data space (i.e., \# of dimension) & 784 & 784\\
\hline
Test accuracy of the chosen model &  94.1& 92.5\\
\hline
\end{tabular}
\end{table}

\subsection{Our proposed framework settings}
In this subsection, we describe the training detail of each stage. Stage (I) utilizes an autoencoder that is pre-trained on the dataset to generate the neighborhood based on the given point-of-interest. Table~\ref{tab2} demonstrates the hyper-parameters of the pre-trained autoencoder for both datasets. Since MNIST contains simpler data points than FashionMNIST, we use a 10-dimensional latent space to represent the images in MNIST, while a 20-dimensional latent space for FashionMNIST.

\begin{table}
\caption{Description of variational autoencoder models used in Stage (I) and classifiers that need to be explained. The model architecture, activation function, and the number of hidden layers are shown accordingly.}\label{tab2}
\begin{tabular}{|l|l|l|}
\hline
 &  VAE & Classifier\\
\hline
MNIST &  \textit{CNV}(32,64, 64), \textit{ReLU}, 10 & \textit{Linear}(20,10), \textit{ReLU}\\
\hline
FashionMNIST &  \textit{CNV}(32,64, 64), \textit{ReLU}, 20 & \textit{Linear}(20,10), \textit{ReLU}\\
\hline
\end{tabular}
\end{table}

\section{Result}
This section will first apply our proposed framework to the MNIST dataset and illustrate how our framework works by providing multiple examples. Then, we apply our method to the FashionMNIST dataset. The examples we presented here demonstrate our framework's potential for improving human understanding of the black-box machine learning models. Note that due to the page limits we only present a handful case studies on two datasets. We also apply our framework on other datasets such as 3-D point cloud data. More interesting examples can be found in our GitHub Page.

\subsection{MNIST}
\begin{figure}[h]
\centering
\includegraphics[width=0.95\textwidth]{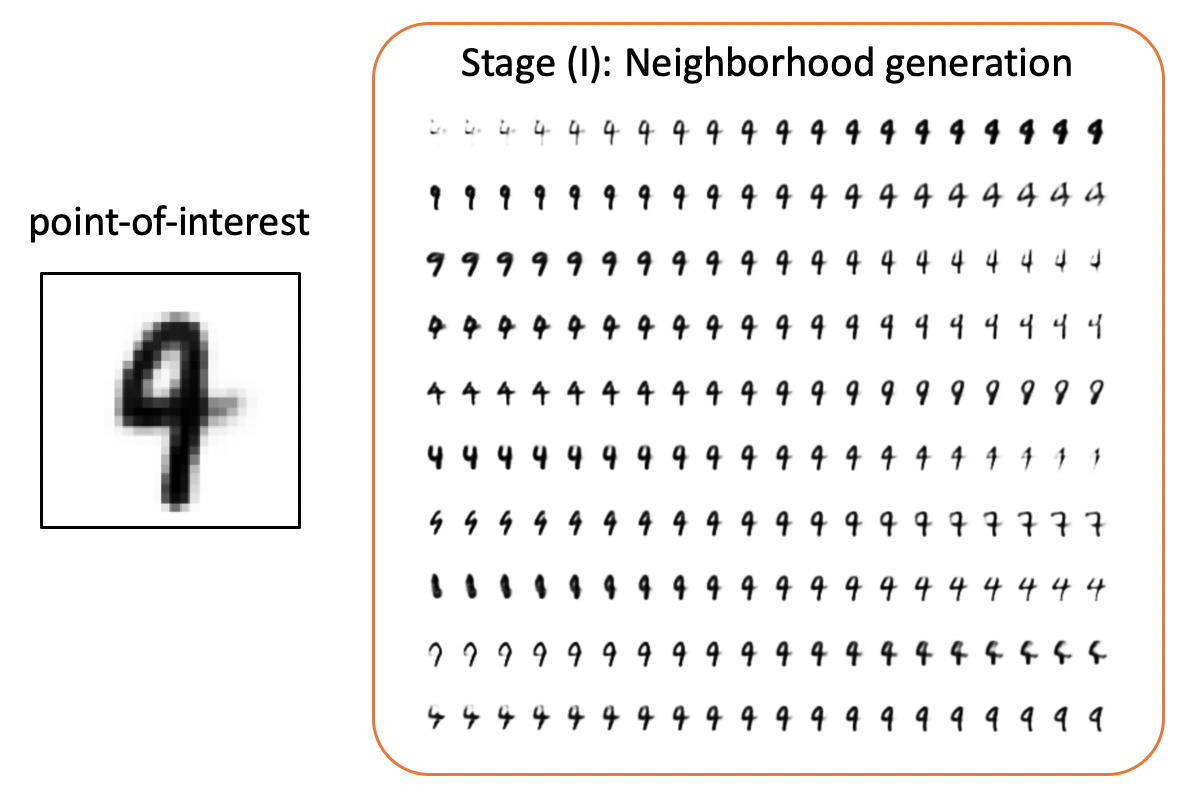}
\caption{Stage (I) of our framework. In Stage (I), the neighborhood of the point-of-interest is generated. The examples in the neighborhood satisfied the criteria in Section 3 as they are all semantically close to the original data point.}
\label{stage1}
\end{figure}

\begin{figure}[h]
\centering
\includegraphics[width=0.95\textwidth]{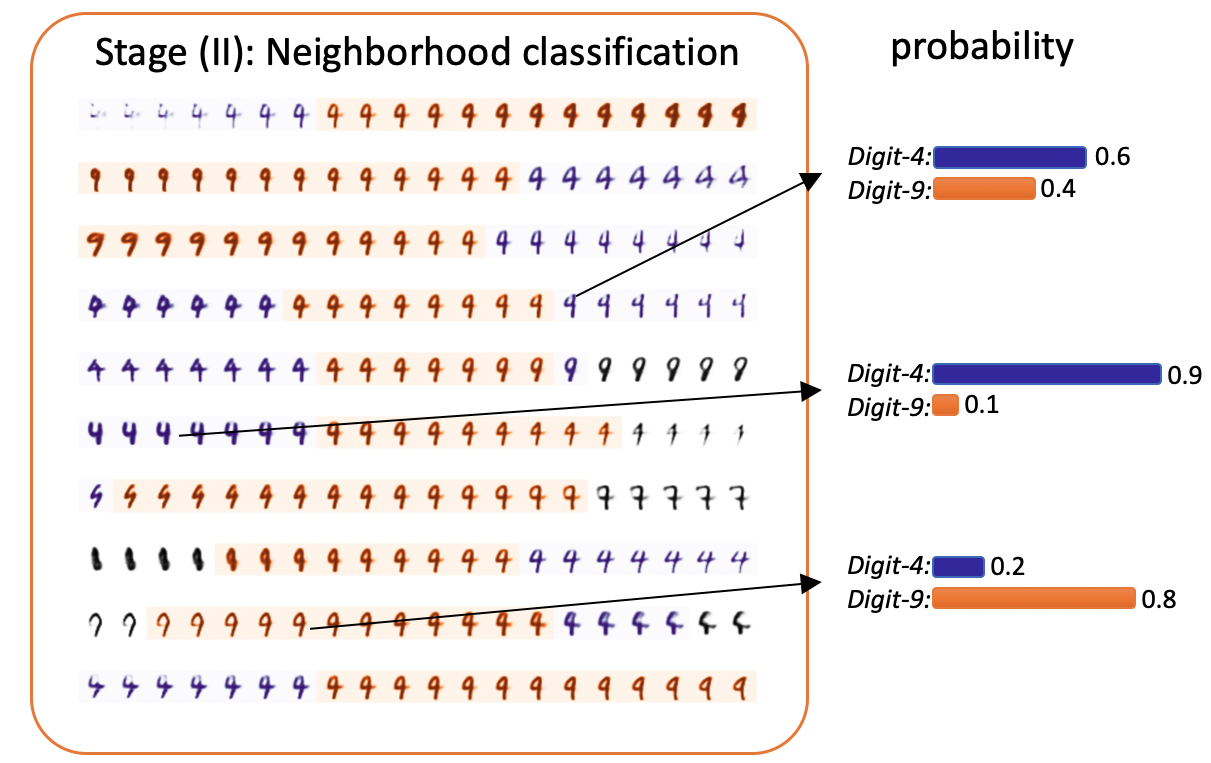}
\caption{Stage (II) of our framework. In Stage (II), the generated neighborhood is classified with the given trained machine learning model. Purple color indicates the image is classified as \textit{digit-4}, orange color indicates the image is classified as \textit{digit-9} and all other classification results are marked as color grey. We also observe that despite being classified to the same label, images close to the decision boundary have higher fidelity.}
\label{stage2}
\end{figure}

A CNN model trained on the MNIST dataset for digit classification is selected and yields a 94.1\% accuracy on the testing dataset. A mispredicted example is chosen for the case study. Figure~\ref{stage1} and Figure~\ref{stage2} show the selected mispredicted point-of-interest and the stage (I) and stage (II) process. As shown in Figure~\ref{stage1}, the neighborhood of the point-of-interest is generated in grey-scale. The examples in the neighborhood satisfied the criteria in Section 3 as they are all semantically close to the original data point. The classified neighborhood is shown in Figure~\ref{stage2}. The colors refer to the classification results. We observe that despite being classified to the same label, images close to the decision boundary have higher fidelity. This observation is consistent with our intuition that the model is more likely mispredicting samples near the decision boundary. One can also draw a similar conclusion by visually examining the classified neighborhood: examples near the decision boundary often have an ambiguous shape that sometimes confuses machine learning models. Through stage (I) and stage (II), our framework generates examples that delineate the model behavior by depicting the local decision boundary.

\begin{figure}[h]
\centering
\includegraphics[width=0.95\textwidth]{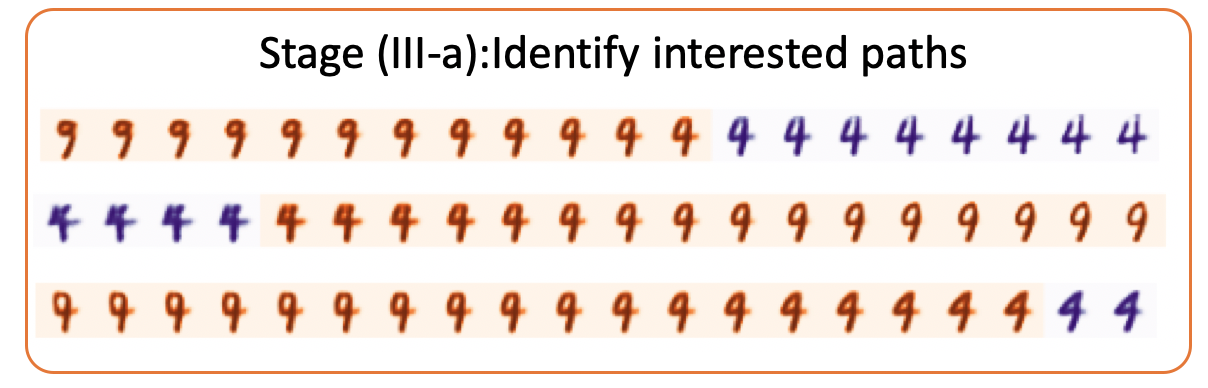}
\caption{Stage (III-a) of the framework. In Stage (III-a), three paths are identified and the morphing is highlighted with different colors. In this example, the three identified morphing paths revealed three related features: 1) the tartness of the circle, 2) the size of the circle and, 3) the straightness of the line.}
\label{stage3a}
\end{figure}

\begin{figure}[h]
\centering
\includegraphics[width=0.95\textwidth]{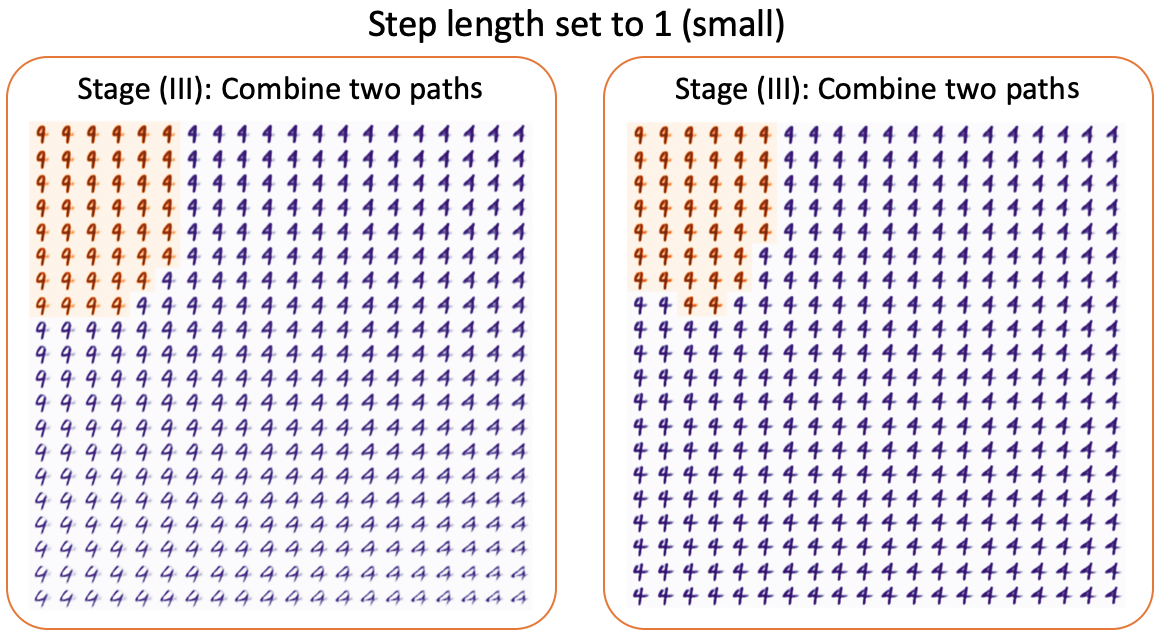}
\caption{Stage (III-b) of the framework. In Stage (III-b), the combination of two paths is presented to achieve a “zoom-in” effect for better carving out the model behavior. From this denser region manifestation, one might conclude that 1) an "open-circle" at the top could help the given predicter correctly identify a \textit{digit-4}, and 2) lines with roundness instead of tartness could mislead the predictor to mispredict a \textit{digit-4} to \textit{digit-9}.}
\label{stage3b_2}
\end{figure}

\begin{figure}[h]
\centering
\includegraphics[width=0.95\textwidth]{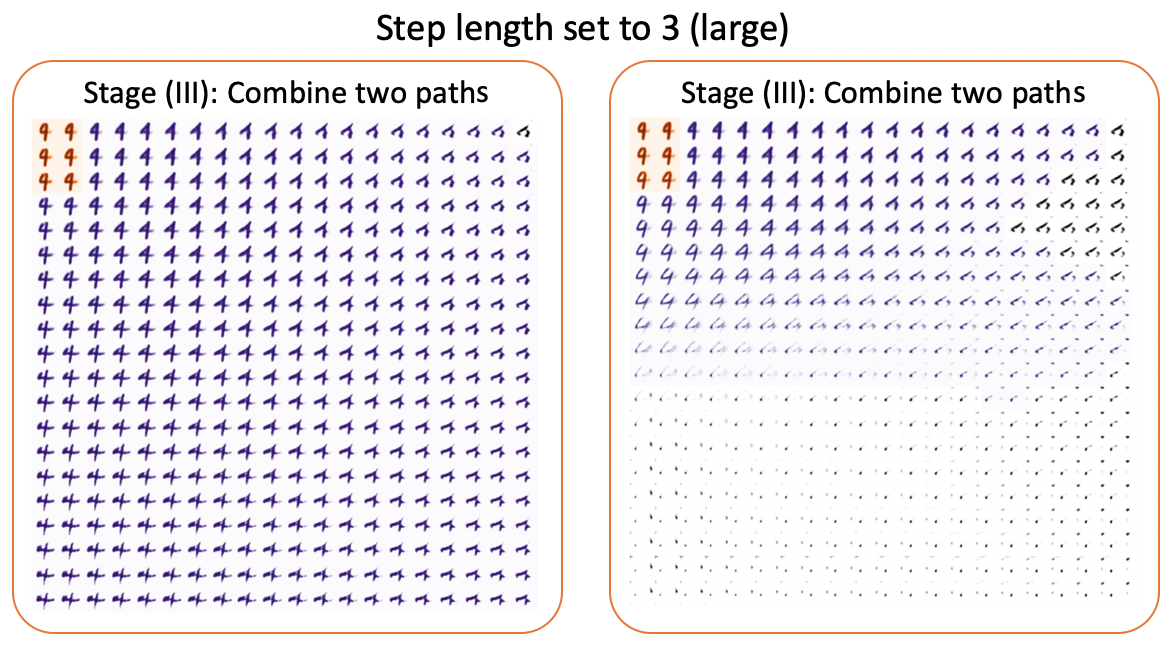}
\caption{Stage (III-b) of the framework. In Stage (III-b), the combination of two paths is presented to achieve a “zoom-out” effect for better carving out the model behavior.}
\label{stage3b_1}
\end{figure}

\begin{figure}[h]
\centering
\includegraphics[width=0.95\textwidth]{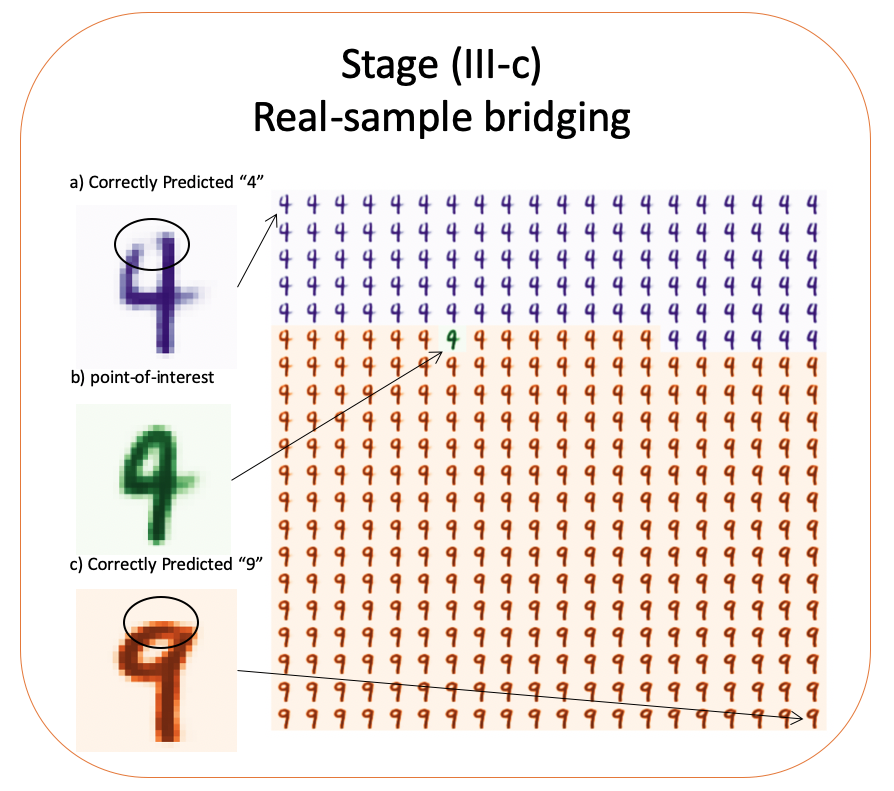}
\caption{Stage (III-c) of the framework. In this stage, two nearest data samples from the original dataset are selected to bridge the gaps between the point-of-interest and real samples on two sides of the decision boundary.}
\label{stage3c}
\end{figure}

After getting the classified neighborhood that carves the local decision boundary around the point-of-interest, human users could be invited to explore the neighborhood using their own intelligence. Figure~\ref{stage3a}, Figure~\ref{stage3b_2}, Figure~\ref{stage3b_1} and Figure~\ref{stage3c} illustrate the three possible human-in-the-loop exploration strategies. From Figure~\ref{stage3a}, one can observe that at stage (III-a) there exist three interesting ways of morphing between \textit{digit-4} and \textit{digit-9}. Therefore, human users can gain insights by investigating the relevant features that have been changed along the process of \textit{digit-4} morphing to \textit{digit-9}. In this example, the three identified morphing paths revealed three related features: 1) the tartness of the circle, 2) the size of the circle and, 3) the straightness of the line. Next, human users can combine two paths for a ``zoom-in''\& ``zoom-out'' investigation. Combining two paths allows human users to gather richer information related to the decision boundary. As shown in Figure~\ref{stage3b_1} and Figure~\ref{stage3b_2}, two possible combinations are chosen and presented, and the \textit{step-length} are adjusted for the ``zoom-in'' effect and the ``zoom-out'' effect. From the denser region manifestation, one might conclude that 1) an "open-circle" at the top could help the given predictor correctly identify a \textit{digit-4}, and 2) lines with roundness instead of tartness could mislead the predictor to mispredict a \textit{digit-4} to \textit{digit-9}. Such conclusions could help human users better understand how the model behaves in a certain region.

Figure~\ref{stage3c} demonstrates the result generated by our third intervention point. As shown in the Figure, a \textit{digit-4} is mispredicted as \textit{digit-9}. By examining the morphing from the nearest digit 4 (in purple) to the nearest digit 9 (in orange), the circled area can be identified by human intelligence as one of the explanations for the misprediction.

\begin{figure}[h]
\centering
\includegraphics[width=0.95\textwidth]{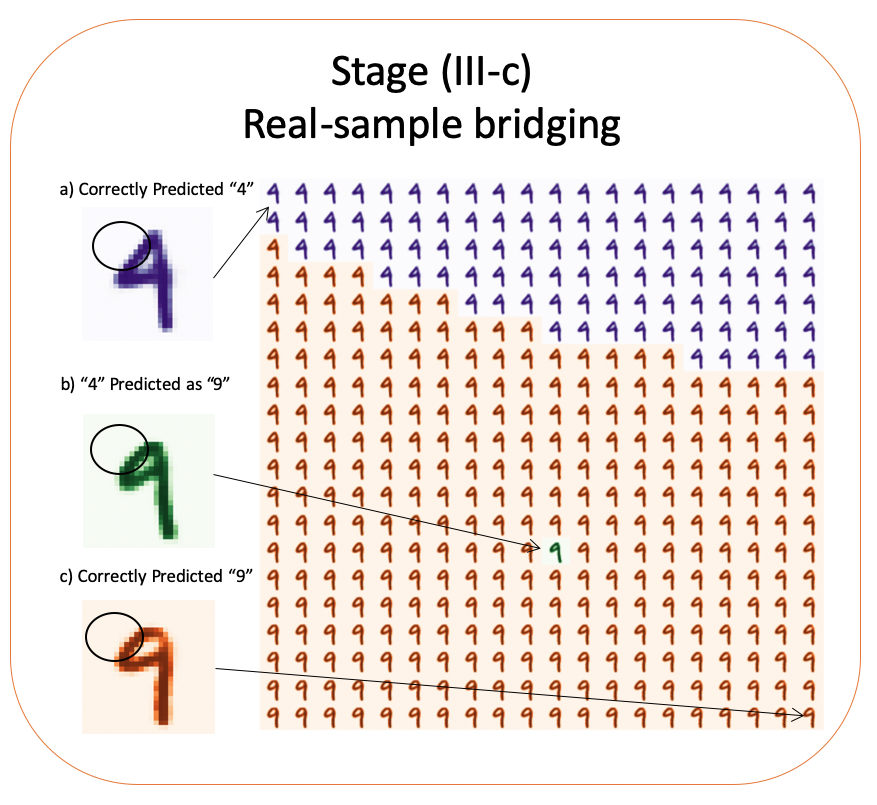}
\caption{Stage (III-c) of the framework. In this stage, two nearest data samples from the original dataset are selected to bridge the gaps between the point-of-interest and real samples on two sides of the decision boundary.}
\label{stage3c_1}
\end{figure}

\begin{figure}[h]
\centering
\includegraphics[width=0.95\textwidth]{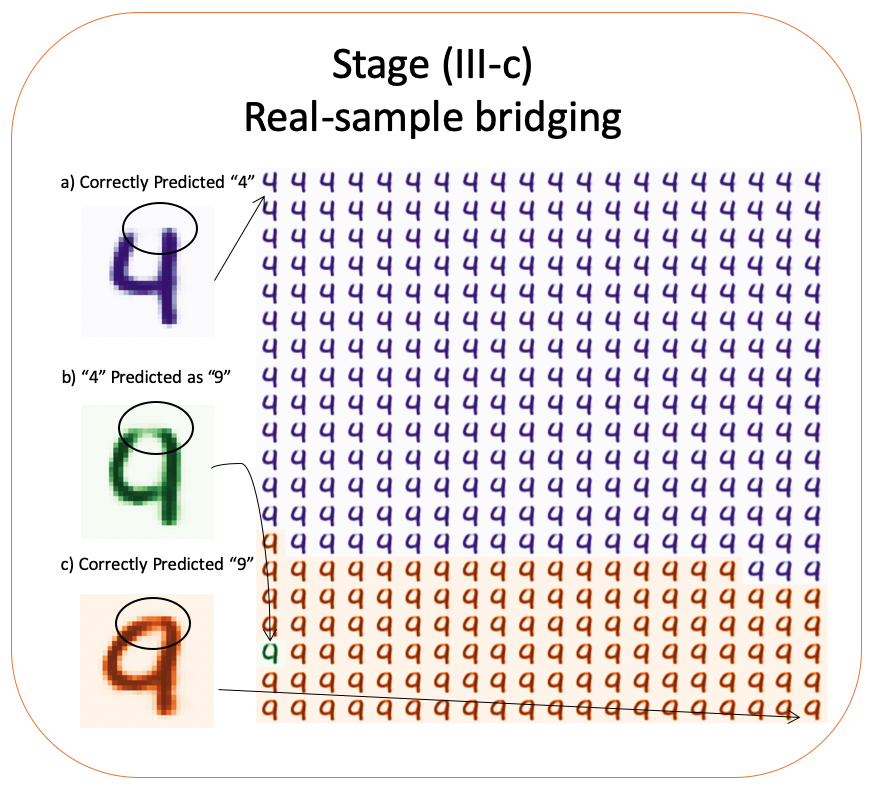}
\caption{Stage (III-c) of the framework. In this stage, two nearest data samples from the original dataset are selected to bridge the gaps between the point-of-interest and real samples on two sides of the decision boundary.}
\label{stage3c_2}
\end{figure}

Two other examples are shown in Figure~\ref{stage3c_1} and Figure~\ref{stage3c_2}. The local decision boundary of the model near the two selected instance-of-interest are displayed, end-users can better understood the model behavior by visually examining these samples. In these two cases, we could observe that the mispredictions are likely to be caused by the circle areas in the image's top-left region. Note that human users can leverage their own intelligence to generate their own understanding with respect to the model behavior. Our framework only provides the intervention points that bridge the gaps between human minds and the black-box nature of machine learning models. 

\subsection{FashionMNIST}
\begin{figure}[h]
\centering
\includegraphics[width=0.95\textwidth]{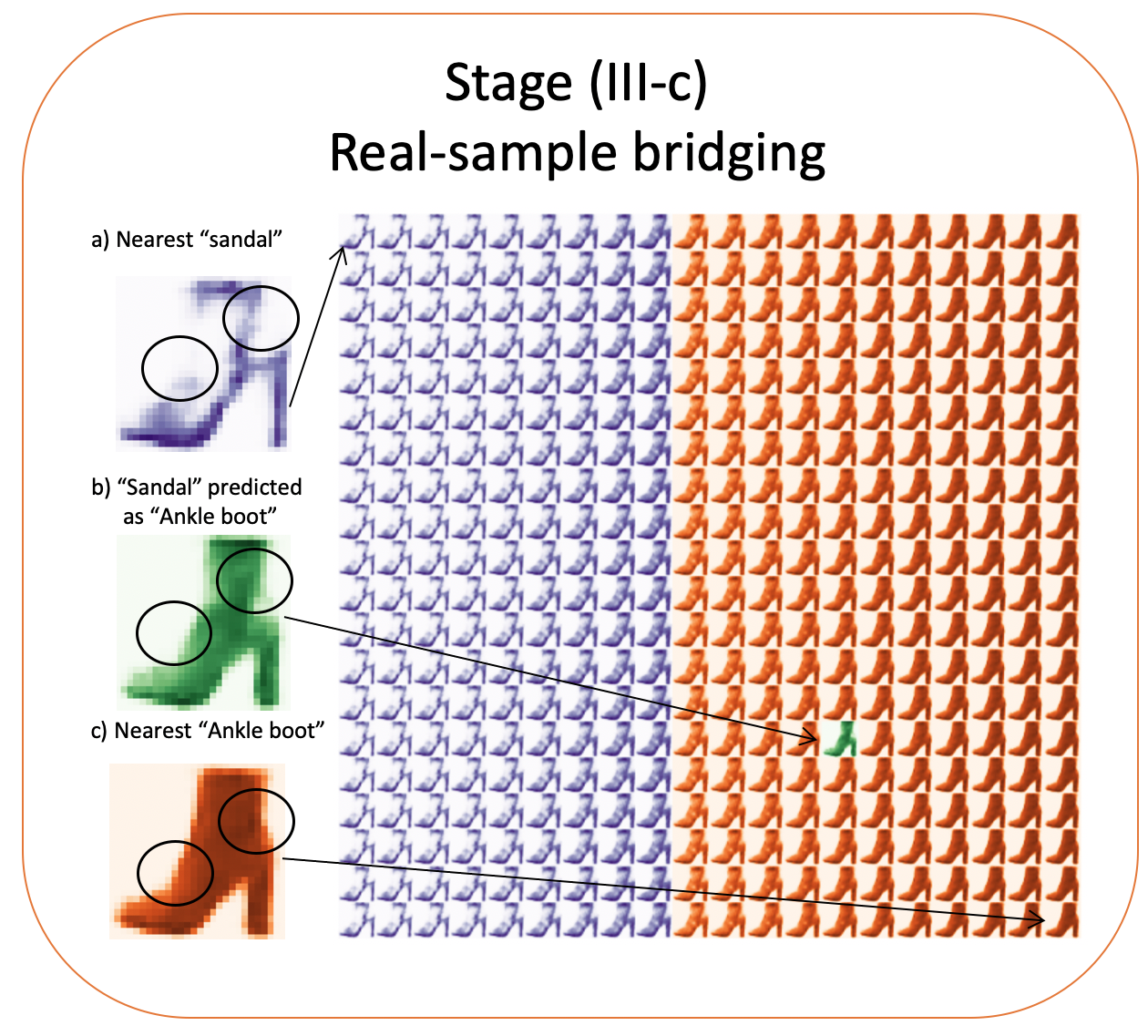}
\caption{A  sandal  is  mispredicted  as  Ankle  boot  in  FashionMNIST dataset.  Without  the  context  that  some  sandal  have  boot-shaped,  it  would  be  difficult  to  understand  the  cause  of  this error. The neighborhood manifestation provided by our framework enable human users to explore the context environment thus gain understanding of this type of mistakes.}
\label{fashionMNIST}
\end{figure}

We provide another experiment using FashionMNIST dataset. In Figure~\ref{fashionMNIST}, a sandal is mispredicted as an ankle boot (in green) by a pre-trained CNN. Without the context that some sandals are boot-shaped, it would be difficult to understand the cause of this error. We select this mispredicted image as an item-of-interest and apply the trained VAE to extract its latent vector. Next, we explore the latent space around the extracted latent vector and generate a manageable number of neighbor images. The trained CNN is then applied to classify the generated images. The decision boundary can be observed as the classified label is morphing from sandal (in purple) to ankle boot (in orange). By visually displaying the neighborhood and the decision boundary (the area that purple turns into orange), the end-user can observe the smooth transition between sandal and ankle boot. Human users can easily draw the conclusion that the circled areas might cause the misprediction, i.e., if a boot-shaped image with blank space at the circled areas, it is likely the image will be classified as ankle boot. 

\section{Workflow of human users of the proposed framework}
This study aims to improve explainability of machine learning models in a human-centric fashion. In this section, we present how a human user or a software engineer can leverage our framework to understand why a given ML model misclassifies a data point. There are three human intervention points.

\subsection{Identifying the point-of-interest}
First, the human user identifies a mispredicted point-of-interest, which software engineers routinely encounter as they debug software systems with ML components. 

\subsection{Identifying interesting dimensions and appropriate step lengths}
Second, the key question from a user's perspective is: how and why a particular region of the point of interest is relevant to the prediction. That is where human users can again contribute by identifying the most interesting dimensions of semantic changes. Our framework leverages a powerful generative model, variational autoencoders, to generate a neighborhood of closely related data points. The generated neighborhood displays a progressive set of plausible variations of the point-of-interest and visualizes the semantic changes across all directions. The human user can use his common sense judgement to identify more interesting dimensions and more appropriate step lengths of changes on these dimensions so that changes in neighboring data-points are perceivable but not too dramatic. 

\subsection{Selecting two most revealing dimensions to generate a matrix for decision boundary visualization}
Third, human users then select two most revealing dimensions so that a matrix of data-points can be generated to visualize the efforts of gradual changes on both dimensions. This matrix represents the neighborhood of interest. All generated data-points in the neighborhood are passed through the actual model-under-investigation so that the decision boundary is identified and visualized verbatim. Human users can gain knowledge and insights by walking through the classified instances and examining the decision boundary. 

These three intervention points provide helpful exploration tools to help human users see, select, and manipulate the neighborhood of the data-point-of-interest and the decision boundary within it and therefore better understand the behavior of the underlying model.

\section{Discussion, Limitations and Future Works}
This paper proposes a human-in-the-loop framework to improve human understanding of the black-box machine learning models locally through verbatim neighborhood manifestation.

However, the proposed method is limited in several ways. First, the neighborhood is generated based on the reconstructed data point. We lack a quantitative measure of the fidelity of the generated neighborhood to the original samples. Though the generated samples are derived from the VAE that was directly trained on the original dataset, some details are lost. Second, we adopt a standard VAE to encode the data point into latent space. Moving in such a latent space typically affects several factors of variation at once, and different directions interfere with each other~\cite{mathieu2019disentangling}. This entanglement effect poses challenges for interpreting these directions' semantic meaning and, therefore, hinders human users from understanding the machine learning models. 

Each of the limitations mentioned above points to a potential direction for future work. We want to quantify the fidelity of the generated data through metrics such as mean-absolute-error or binary-cross-entropy. For the second limitation, we are considering leveraging disentangle-VAE to generate neighborhoods along with semantic meaningful directions. We are also interested in learning a set of latent space directions inducing orthogonal transformations that are easy to distinguish from each other and offer robust semantic manipulations in the neighborhood manifestation. These future works introduce exciting challenges for bridging the gaps between the black-box nature of machine learning models and human understanding. 

\section{Conclusion}
Machine learning models are mainly being developed and fine-tuned for optimal accuracy, while understanding these models has not attracted much attention. Existing XAI models focus on providing approximate hit-or-miss explanations, which do not involve humans in explaining and neglect human intelligence. We propose a human-in-the-loop explanation framework that reframes the explanation problem as a human-interactive problem to tackle this limitation. Our approach utilizes a generative model to enrich the (mispredicted) point-of-interest neighborhood and crave out the local decision boundary by highlighting the model prediction results. We provide three human-involved exploration intervention points that assist human users to leverage their own understanding of the model behavior. We conducted case studies on two datasets, and the experimental results demonstrate the potential of our framework for building a bridge between machine and human intelligence.

%
%
%
%

\bibliographystyle{splncs04}
\bibliography{reference}

\begin{thebibliography}{10}
\providecommand{\url}[1]{\texttt{#1}}
\providecommand{\urlprefix}{URL }
\providecommand{\doi}[1]{https://doi.org/#1}

\bibitem{ancona2017towards}
Ancona, M., Ceolini, E., {\"O}ztireli, C., Gross, M.: Towards better
  understanding of gradient-based attribution methods for deep neural networks.
  arXiv preprint arXiv:1711.06104  (2017)

\bibitem{ancona2019explaining}
Ancona, M., {\"O}ztireli, C., Gross, M.: Explaining deep neural networks with a
  polynomial time algorithm for shapley values approximation. arXiv preprint
  arXiv:1903.10992  (2019)

\bibitem{bach2015pixel}
Bach, S., Binder, A., Montavon, G., Klauschen, F., M{\"u}ller, K.R., Samek, W.:
  On pixel-wise explanations for non-linear classifier decisions by layer-wise
  relevance propagation. PloS one  \textbf{10}(7),  e0130140 (2015)

\bibitem{chattopadhay2018grad}
Chattopadhay, A., Sarkar, A., Howlader, P., Balasubramanian, V.N.: Grad-cam++:
  Generalized gradient-based visual explanations for deep convolutional
  networks. In: 2018 IEEE Winter Conference on Applications of Computer Vision
  (WACV). pp. 839--847. IEEE (2018)

\bibitem{kabra2015understanding}
Kabra, M., Robie, A., Branson, K.: Understanding classifier errors by examining
  influential neighbors. In: Proceedings of the IEEE conference on computer
  vision and pattern recognition. pp. 3917--3925 (2015)

\bibitem{karimi2019characterizing}
Karimi, H., Derr, T., Tang, J.: Characterizing the decision boundary of deep
  neural networks. arXiv preprint arXiv:1912.11460  (2019)

\bibitem{karimi2020decision}
Karimi, H., Tang, J.: Decision boundary of deep neural networks: Challenges and
  opportunities. In: Proceedings of the 13th International Conference on Web
  Search and Data Mining. pp. 919--920 (2020)

\bibitem{mathieu2019disentangling}
Mathieu, E., Rainforth, T., Siddharth, N., Teh, Y.W.: Disentangling
  disentanglement in variational autoencoders. In: International Conference on
  Machine Learning. pp. 4402--4412. PMLR (2019)

\bibitem{lime}
Ribeiro, M.T., Singh, S., Guestrin, C.: "why should {I} trust you?": Explaining
  the predictions of any classifier. In: Proceedings of the 22nd {ACM} {SIGKDD}
  International Conference on Knowledge Discovery and Data Mining, San
  Francisco, CA, USA, August 13-17, 2016. pp. 1135--1144 (2016)

\bibitem{rudin2019stop}
Rudin, C.: Stop explaining black box machine learning models for high stakes
  decisions and use interpretable models instead. Nature Machine Intelligence
  \textbf{1}(5),  206--215 (2019)

\bibitem{selvaraju2017grad}
Selvaraju, R.R., Cogswell, M., Das, A., Vedantam, R., Parikh, D., Batra, D.:
  Grad-cam: Visual explanations from deep networks via gradient-based
  localization. In: Proceedings of the IEEE international conference on
  computer vision. pp. 618--626 (2017)

\bibitem{shrikumar2017learning}
Shrikumar, A., Greenside, P., Kundaje, A.: Learning important features through
  propagating activation differences. arXiv preprint arXiv:1704.02685  (2017)

\bibitem{shrikumar2016not}
Shrikumar, A., Greenside, P., Shcherbina, A., Kundaje, A.: Not just a black
  box: Learning important features through propagating activation differences.
  arXiv preprint arXiv:1605.01713  (2016)

\bibitem{Simonyan2014DeepIC}
Simonyan, K., Vedaldi, A., Zisserman, A.: Deep inside convolutional networks:
  Visualising image classification models and saliency maps. CoRR
  \textbf{abs/1312.6034} (2014)

\bibitem{sundararajan2017axiomatic}
Sundararajan, M., Taly, A., Yan, Q.: Axiomatic attribution for deep networks.
  arXiv preprint arXiv:1703.01365  (2017)

\bibitem{wang2020score}
Wang, H., Wang, Z., Du, M., Yang, F., Zhang, Z., Ding, S., Mardziel, P., Hu,
  X.: Score-cam: Score-weighted visual explanations for convolutional neural
  networks. In: Proceedings of the IEEE/CVF Conference on Computer Vision and
  Pattern Recognition Workshops. pp. 24--25 (2020)

\bibitem{wang2019deepvid}
Wang, J., Gou, L., Zhang, W., Yang, H., Shen, H.W.: Deepvid: Deep visual
  interpretation and diagnosis for image classifiers via knowledge
  distillation. IEEE transactions on visualization and computer graphics
  \textbf{25}(6),  2168--2180 (2019)

\bibitem{zeiler2014visualizing}
Zeiler, M.D., Fergus, R.: Visualizing and understanding convolutional networks.
  In: European conference on computer vision. pp. 818--833. Springer (2014)

\end{thebibliography}
\end{document}